\documentclass{bmvc2k}

\title{Sparse in Space and Time: Audio-visual Synchronisation with Trainable Selectors}

\addauthor{Vladimir Iashin}{vladimir.iashin@tuni.fi}{1}
\addauthor{Weidi Xie}{weidi@robots.ox.ac.uk}{2,3}
\addauthor{Esa Rahtu}{esa.rahtu@tuni.fi}{1}
\addauthor{Andrew Zisserman}{az@robots.ox.ac.uk}{3}

\addinstitution{
 Computing Sciences\\
 Tampere University\\
 Tampere, Finland
}
\addinstitution{
 Coop. Medianet Innovation Center\\
 Shanghai Jiao Tong University \\
 Shanghai, China
}
\addinstitution{
 Visual Geometry Group\\
 Department of Engineering Science\\
 University of Oxford\\
 Oxford, UK
}

\runninghead{Iashin, Xie, Rahtu, Zisserman}{Audio-visual Synchronisation}

\def\eg{\emph{e.g}\bmvaOneDot}
\def\ie{\emph{i.e}\bmvaOneDot}

\def\etal{\emph{et al}\bmvaOneDot}

\definecolor{brick}{HTML}{B85450}
\newcommand{\blue}[1]{#1}

\newcommand{\comma}{, \, }

\usepackage{wrapfig} 
\usepackage{booktabs} 
\usepackage{amsfonts}
\usepackage{amssymb}
\usepackage{pifont}
\usepackage{capt-of}
\begin{document}

\maketitle

\vspace{-4ex}
\begin{abstract}
The objective of this paper is audio-visual synchronisation of general videos `in the wild'.
For such videos, the events that may be harnessed for synchronisation cues may be spatially small and may occur
only infrequently during a many seconds-long video clip, \ie the synchronisation signal is `sparse in space and time'.
This contrasts with the case of synchronising videos of talking heads, where audio-visual correspondence is dense
in both time and space.

We make four contributions: 
\textit{(i)} in order to handle longer temporal sequences required for sparse synchronisation signals,
we design a multi-modal transformer model that employs `selectors' to 
distil the long audio and visual streams into small sequences that are then used to predict the temporal offset between streams.
\textit{(ii)} We identify  artefacts that can arise from the compression codecs used for audio and video and can be
used by audio-visual models in training to artificially solve the synchronisation task. 
\textit{(iii)}~We curate a dataset with only sparse in time and space synchronisation signals; and  
\textit{(iv)}~the effectiveness of the proposed model is shown on both dense and sparse datasets quantitatively and qualitatively.
Project page: \href{https://v-iashin.github.io/SparseSync}{\color{blue} \texttt{\textbf{v-iashin.github.io/SparseSync}}}

\end{abstract}

\section{Introduction}
\label{sec:intro}
Audio-visual synchronisation is the task of determining the temporal offset between the audio (sound) and visual (image) streams in a video.
In recent literature, this task has been explored by exploiting strong correlations between the audio and visual streams, 
\eg in human speech~\cite{Chung16a,Chung2019perfect,Afouras20b} and
playing instruments~\cite{arandjelovic2018objects,Owens18audio-visual}, to provide a training signal for deep neural networks.
In such scenarios, effective signals for synchronisation can be discovered between the lip or body movements and audio at almost every second.
Despite the tremendous success achieved by these methods, for the most part, existing models are still limited to specialised domains, 
and not directly applicable to general (non-face, non-music) classes.

\begin{figure}[h]
    \centering
    \vspace{-1ex}
    \includegraphics[width=1.0\textwidth]{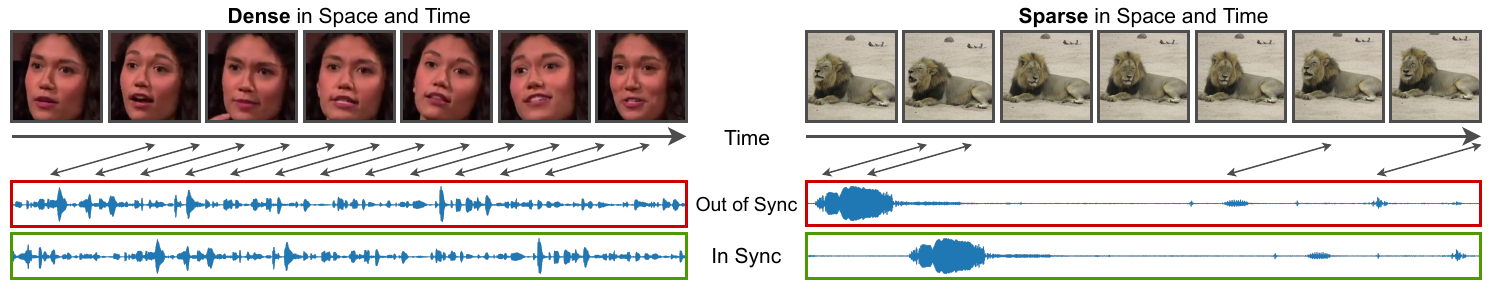}
    \vspace{-4ex}
    \caption{\normalsize
        Audio-visual synchronisation requires a model to relate changes in the visual  and audio streams. 
        Open-domain videos often have a small visual indication, \ie sparse in space.
        Moreover, cues may be intermittent and scattered, \ie sparse across time, \eg a lion only roars once during a video clip. This differs from a tight face crop of a speaker where cues are dense in space and time.
    }
    \label{fig:teaser}
    \vspace{-18pt}
\end{figure}

Our goal in this paper is to develop the next-generation audio-visual (AV) synchroniser.
Rather than focusing on a specialised domain, such as human speech, 
we explore architectures for AV synchronisation for videos of general thematic content,
{\em e.g.}~daily videos~\cite{Korbar18, Khosravan19, Chen20a} and live sports~\cite{Ebeneze21-tennis}.
A solution for this task would be extremely useful for a number of applications that improve a user's viewing 
experience -- in order to avoid or at least automatically detect AV synchronisation offsets. 
Applications such as video conferencing, television broadcasts, and video editing, are currently largely done 
by `off-line' measurements or heavy manual processing~\cite{Shrestha10,Staelens12,Dassani19}. 

However, upgrading the existing audio-visual synchronisation systems to general videos is not straightforward, due to the following challenges:
{\em (i)}, in general videos, the synchronisation signal is often {\em sparse and instantaneous in time}, 
(a lion roaring or a tennis volley), rather than {\em dense in time} (a recorded monologue);
{\em (ii)}, objects that emit sounds can vary in size or appear in the distance making their presence on the frame 
 small or {\em sparse in space} (a ball being hit in tennis), 
whereas synchronisation of a talking-head video may rely on visual cues from the localised mouth region, \ie \textit{dense in space};
{\em (iii)}, some sound sources do not have a useful visual signal for synchronisation,~\eg~stationary sounds (a car 
engine or electric trimmer), ambient sounds (wind, water, crowds, or traffic), and off-screen
distractors (commentary track or advertisements);
{\em (iv)}, video encoding algorithms compress unperceived redundancy of a signal,
this, however, can introduce artefacts that may lead to a trivial solution when training for audio-visual synchronisation; 
lastly, {\em(v)} due to its challenging nature, a public benchmark to measure progress has not yet been established.

In this paper, 
\textit{(i)} we introduce a novel multi-modal transformer architecture, \textbf{SparseSelector}, that can digest long videos with
linear scaling complexity with respect to the number of input tokens,
and predict the temporal offset between the audio and visual streams. 
We achieve this by using a set of learnable {\em queries} to select informative signals from the `sparse' video events 
across a wide time span.
\textit{(ii)} We show that for specific common audio and visual coding standards,
a model can detect compression artefacts during training.
We present a few simple indicators to determine if a model has learnt using these artefacts, 
as well as suggest several ways to mitigate the problem.
Specifically, for the RGB stream, we recommend avoiding the MPEG-4 Part 2 codec, as well as reducing 
the sampling rate for audio.
\textit{(iii)} Additionally, to measure the progress of audio-visual synchronisation on general thematic content, 
we curate a subset of VGGSound with `sparse' audio-visual correspondence called VGGSound-Sparse. 
We validate the effectiveness of the new model with thorough experiments on the existing lip reading benchmark~(LRS3) and
natural videos from VGGSound-Sparse and demonstrate state-of-the-art performance.

\section{Related Work}
\paragraph{Audio-visual synchronisation.}
During the pre-deep-learning era, the audio-visual human face synchronisation models relied on manually
crafted features and statistical models \cite{hershey1999audio,slaney2000facesync}. With the advent of deep learning, \cite{chung2016lip} introduced a
two-stream architecture that was trained in a self-supervised manner using a binary contrastive loss. Later improvements were brought by multi-way contrastive training~\cite{Chung2019perfect}, and Dynamic Time Warping \cite{rabiner1993fundamentals} used by \cite{halperin2019dynamic}.
Khosravan~\etal~\cite{Khosravan19} demonstrated the benefits
of spatio-temporal attention and Kim \etal \cite{kim2021end} employed a cross-modal embedding matrix to
predict the offset for synchronisation. 
The progress was followed by \cite{kadandale2022vocalist} who introduced an architecture called VocaLiST 
with three transformer decoders~\cite{vaswani2017attention}: two that cross-attend individual modalities and a third that fuses the outputs of the first two. 
These methods achieve impressive performance but focus on human speech rather than open-domain videos.

Although audio-visual synchronisation of general classes is a novel task, a few promising attempts have been made. 
In particular, Casanovas \etal \cite{casanovas2015audio} studied a handful of different scenes captured from a set of cameras. 
More recently, Chen \etal \cite{chen2021audio} adapted the transformer architecture
and used a subset of VGGSound \cite{Chen20a} covering 160 classes.
In contrast to prior work, we focus on more challenging classes that have `sparse' rather than `dense' synchronisation signals. 

\paragraph{Video coding artefacts.}
Since the early work of Doersch \etal on self-supervision~\cite{doersch2015unsupervised},
it has been known that network training can find shortcuts. Similarly, shortcuts due to 
video editing and coding artefacts have been noted in
Wei \etal \cite{wei2018learning} and Arandjelović \etal \cite{arandjelovic2018objects}.
In particular, \cite{wei2018learning} tackled the arrow-of-time in videos and studied artificial cues 
caused by black regions on video frames.
While~\cite{arandjelovic2018objects} noticed a slight impact of MPEG-encoding on audio-visual correspondence 
training and attributed it to the way negative samples are picked with respect to the start time of a positive sample.
In this work, we study the ways to easily spot that the data contains artificial signals, as well as provide a 
few recommendations on how to prevent leaking such artefacts into data.

\section{\textbf{SparseSelector}: an Audio-visual Synchronisation Model}
\label{sec:architecture}

In this section, we describe our audio-visual synchronisation model, 
where the audio-visual correspondence may only be  available at sparse events in the `in the wild'  videos.
This requires the model to handle longer video clips so that there is a high probability that a synchronisation event will occur. 
To this end, we propose \textit{SparseSelector}, 
a transformer-based architecture that enables the processing of long videos with linear complexity with 
respect to the duration of a video clip.
It achieves this by `compressing' the audio and visual input tokens into two small sets of learnable \textit{selectors}. 
These selectors form an input to a transformer which predicts the temporal offset between the audio and visual streams.

\begin{figure}[t]
    \centering
    \includegraphics[width=\textwidth]{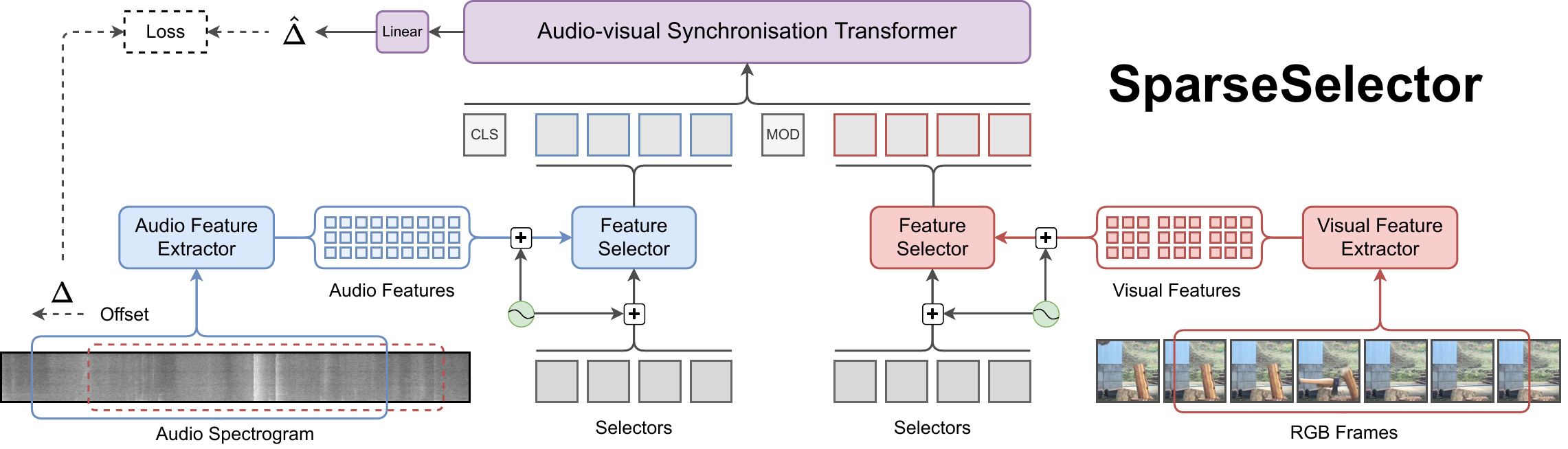}
    \caption{\normalsize 
        \textbf{An overview of \textbf{SparseSelector}}. 
        The input is a spectrogram of the audio waveform and RGB frames from the video stream. These are passed through corresponding feature extractors, and 
        the resulting features are refined with trainable selectors that `pick' useful cues for synchronisation.
        As a result, the synchronisation transformer operates on substantially shorter sequences than the original input.
        The visual and audio selector queries are concatenated with classification (\texttt{CLS}) 
        and separation tokens (\texttt{MOD}) as input to the transformer.
        Finally, the CLS token of the transformer output  is used to predict the audio offset using a linear classification head.
        RGB frames are zoomed-in for visualisation purposes. The model is trained by off-setting the audio spectrogram. 
        Dashed lines illustrate train-time behaviour.
    }
    \label{fig:arch}
    \vspace{-10pt}
\end{figure}

\paragraph{Architecture overview.}
The overview of the model is shown in Fig.~\ref{fig:arch}.
Given 
an audio spectrogram 
$\mathcal{A} \in \mathbb{R}^{H_a \times W_a \times 1}$ \blue{($H_a$, $W_a$ are frequency and time dimensions)} 
and 
a stack of RGB frames 
$\mathcal{V} \in \mathbb{R}^{T_v \times H_v \times W_v \times 3}$,
the audio-visual synchronisation model outputs the offset $\Delta$ between 
audio ($\mathcal{A}$) and visual ($\mathcal{V}$) streams:
\begin{align}\label{eq:sync_main}
    \Delta = 
    \Phi_{\text{Sync}}\Big(\Phi_{\text{A-Sel}}\big(\Phi_{\text{A-Feat}}(\mathcal{A})\big)\comma 
    \Phi_{\text{V-Sel}}\big(\Phi_{\text{V-Feat}}(\mathcal{V})\big)\Big). 
\end{align}
First, audio and visual streams are independently encoded in feature extraction modules $\Phi_\text{A/V-Feat}$.
Next, trainable \textit{selectors} are passed to \textit{Feature Selectors} ($\Phi_\text{A/V-Sel}$) along 
with the encoded features where they `summarise' informative signals from the features that contain `sparse' 
information across time and space.
Finally, the selectors are used in \textit{Synchronisation Transformer} ($\Phi_\text{Sync}$)
to predict the temporal offset $\Delta$ between audio and visual streams.\\[-0.8cm]

\paragraph{Feature encoding.}
\!\!\!Audio \& visual inputs are encoded in spatio-temporal~feature extractors: 
\begin{align}
    a = \Phi_{\text{A-Feat}}(\mathcal{A}) \in \mathbb{R}^{h_a \times w_a \times d_a}, \hspace{10pt}
    v = \Phi_{\text{V-Feat}}(\mathcal{V}) \in \mathbb{R}^{t_v \times h_v \times w_v \times d_v},
\end{align}
where $t\comma h\comma w$, and $d$ denote time, height, width and channel dimensions, respectively.
For the audio backbone, we use a variant of ResNet18 \cite{he2016deep}, 
which we pre-train on VGGSound \cite{Chen20a} for sound classification.
As for the visual backbone, we adopt S3D \cite{xie2018rethinking} pre-trained for action recognition 
on Kinetics 400 \cite{kay2017kinetics}.
Although the setting allows employing any visual recognition network, 
we found that training a synchronisation model with a frame-wise feature extractor was significantly 
more difficult.\\[-0.8cm]

\paragraph{Feature Selectors.}
\blue{To utilize sparsely occurring synchronisation cues,
the model should be able to handle longer input sequences.}
Moreover, accurate synchronisation requires a higher visual frame rate than other video understanding
tasks (\eg action recognition), which further increases the input size. 
For this reason, drawing on the idea of trainable queries~\cite{carion2020end,jaegle2021perceiver,Xu2021},
we propose to use a small number of trainable `selectors' that learn to attend to the most useful modality features for 
synchronisation and, thus, reducing sequence length.

The architecture of the \textit{Feature Selector} is similar to the transformer decoder \cite{vaswani2017attention}.
Specifically, we start by flattening audio and visual features into sequences 
$a \in \mathbb{R}^{h_a w_a \times d_a}$ and 
$v \in \mathbb{R}^{t_v h_v w_v \times d_v}$.
After adding trainable positional encoding ($\text{PE}_*$) for each dimension, 
trainable selectors and modality features are passed to the separate Feature Selectors as follows:
\begin{align}
    \hat{q}_a = \Phi_{\text{A-Sel}}(a+\text{PE}_{a}, q_a + \text{PE}_{q_a}), \hspace{10pt}
    \hat{q}_v = \Phi_{\text{V-Sel}}(v+\text{PE}_{v}, q_v + \text{PE}_{q_v}), 
\end{align}
where $q_a,\hat{q}_a \in \mathbb{R}^{k_a \times d}$ and $q_v,\hat{q}_v \in \mathbb{R}^{k_v \times d}$ 
while $k_\text{a/v}$ are the numbers of selectors.

Note that the selectors provide a `short summary' of the context features through the cross-attention mechanism 
while making the memory footprint more manageable.
The reduced memory requirement is a consequence of (a) casting the complexity from quadratic to linear 
w.r.t.\ the input length, and (b) setting $k_v \ll t_v\cdot h_v\cdot w_v $ and $k_a \ll h_a\cdot w_a$.
The number of selectors ($k_{v}$ or $k_{a}$) can be conveniently tweaked according to the memory budget.\\[-0.8cm]

\paragraph{Audio-visual synchronisation transformer.}
To fuse the audio-visual cues from individual selectors,
we adopt the standard transformer encoder layers to jointly process them,
and to predict the offset, 
{\em i.e.}~relative temporal shift between audio and visual streams, as follows:
\begin{align}\label{eq:sync_late}
    \hat{\Delta} = \Phi_{\text{Sync}}([\texttt{CLS}; \text{ } q_v; \text{ } \texttt{MOD}; \text{ } q_a])
\end{align}
Here we concatenate the visual-audio selectors with two learnable special tokens, namely the classification 
token~$[\texttt{CLS}]$,
and the modality token~$[\texttt{MOD}]$ that separates the two modalities. 
The offset prediction is obtained by applying a linear prediction head on the first token of the output sequence 
(omitted from Eq.~\eqref{eq:sync_main} and \eqref{eq:sync_late} for clarity).\\[-0.8cm]

\paragraph{Training procedure.}
We assume that the majority of videos in the public datasets are synchronised to a good extent. 
With this assumption, we can artificially create temporal offsets between audio and visual streams from a video.
We formulate the audio-visual synchronisation as a classification task onto a set of offsets 
from a pre-defined temporal grid space as 
$[-2.0\comma -1.8\comma \dots\comma 0.0\comma +0.2\comma \dots\comma +2.0]$ sec.
The step size is motivated by the $\pm$0.2 sec human tolerance,
where the ITU performed strictly controlled tests with expert viewers and found that the threshold for acceptability is $-$0.19sec to $+$0.1sec~\cite{radiocommunicationrelative}.
To train the model, we employ the cross-entropy loss.
For our experiments, we randomly trim a 5-sec segment out of 9 seconds such that both audio and visual streams are within the 
9-second clip to make inputs of the same size and avoid padding that could hint if the input is off-sync.

\section{Avoiding Temporal Artefacts}
\vspace{-1ex}

In this section, we describe our discovery of trivial solutions for training audio-visual synchronisation,
that is, the model is able to exploit the video compression artefacts, to infer the time stamp for the specific video clip.
Additionally, we also detail a suite of techniques that allows us to probe the artefacts and provide 
some practical suggestions to avoid them.

\vspace{-1ex}
\subsection{Identifying Temporal Artefact Leakage}\label{sec:artefacts}

\begin{table}[t]
    \centering
    \begin{minipage}[c]{0.4\textwidth}
        \centering
        \includegraphics[width=0.9\textwidth]{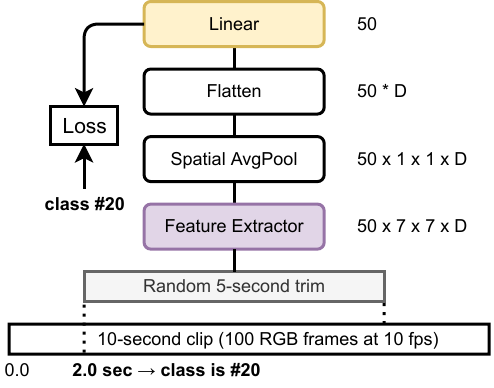}
    \end{minipage}\hfill
    \begin{minipage}[c]{0.6\textwidth}
        \centering 
        \small
        \vspace{-1ex}
        \setlength\tabcolsep{0.2em}
        \begin{tabular}{l r r}
            \toprule
            \textbf{Codec} & \textbf{Acc@1} & \textbf{Acc@5} \\
            \midrule
            MPEG-4 Part 2 (\texttt{mpeg4}) & 27.2 & 77.1 \\
            MPEG-4 Part 10 (H.264) & 2.5 & 11.9 \\
            ProRes & 2.7 & 13.4 \\
            \midrule
            AAC @ 44100Hz & 86.7 & 100.0 \\
            AAC @ 22050Hz & 23.0 & 74.3 \\
            AAC @ 16000Hz & 6.3 & 19.3 \\
            Lossless @ 22050Hz & 2.9 & 14.6 \\
            \bottomrule
        \end{tabular} 
    \end{minipage}
    \caption{\normalsize 
        \textbf{Commonly used coding standards may leak temporal artefacts -- it is easy to test}. 
        \textit{Left}: a simple architecture trained to classify the start of a trim from a 10s clip to a 
        pre-defined 0.1 sec-step grid (here for RGB). 
        \textit{Right}: Accuracy comparison for RGB and audio stream codecs predicting the start of an RGB or audio trim. 
        Metrics are accuracy at 1 and 5 on 50 classes. Chance performance is 2 and 10\,\%. 
        The higher accuracies indicate that an artefact is being used -- see text for discussion.
    }
    \vspace{-3ex}
    \label{tab:start_pred_arch}
\end{table}

We present two ways of identifying the temporal artefact leakage.
In particular, training to predict the start time of a temporal crop (discussed next) and 
tracking metrics with temporal tolerance (discussed in the supplementary material).

\vspace{-2ex}
\paragraph{Training to predict a video clip's time stamp.}
A synchronisation model should rely solely on temporal positions of conceptual cues instead of, what we call, 
temporal artefacts.
To check if data is polluted with artefacts, we suggest training a model to predict the start time of a random trim of an
available video clip as shown in Tab.~\ref{tab:start_pred_arch}, left.
Of course, it should not be possible to determine the start time of the trim in the original clip from the trim itself, 
and a network trained for this task should achieve only chance performance.  
However, for some audio and video codecs, the performance is far higher indicating artefact leakage. 

The start-time classifier is a  simple feature extractor (ResNet18). It is trained on three variations
 of the MJPEG-AoT dataset \cite{wei2018learning} obtained from the Vimeo streaming service: 
original ProRes videos,  and ProRes videos transcoded into either MPEG-4 Pt.\,2 
(aka.\,\texttt{mpeg4}) or MPEG-4 Pt.\,10 (aka.\,H.264). 
Note, frames in ProRes are compressed independently from others.
If the visual stream of the video is encoded using \texttt{mpeg4}, 
the model trained to predict the start of the trim can do it significantly beyond a chance performance 
(Tab.~\ref{tab:start_pred_arch}, top-right). 

Similarly, Advanced Audio Coding (AAC) might also leak temporal cues to the audio signal (Tab.~\ref{tab:start_pred_arch}, bottom-right).
Since it is challenging to find a large set of videos with lossless audio compression, we used audio of randomly generated noise
with a specified sampling rate
and saved it to a disk losslessly (PCM) to obtain the performance with lossless compression. 
To obtain results on AAC, we transcoded these files to AAC with \texttt{ffmpeg}.

\subsection{Preventing Temporal Artefact Leakage}\label{sec:prev_temp_leakage}
\paragraph{Avoiding MPEG-4 Part 2 in favour of H.264.} 
The algorithm that selects key-frames in MPEG-4 Part 2 (\texttt{mpeg4}) is less flexible than the one of H.264.
In particular, \texttt{ffmpeg}, which is commonly used in practice, by default, encodes key frames every 12 frames.
This means that each of the following 11 frames is merely a residual of the key frame and it is noticeable on the RGB stream
(as we show in the supplementary).
Such a temporal regularity can be picked up by a model and used to solve the task relying mostly on these artefacts.
In contrast, each frame encoded by H.264 can reference up to 16 key-frames,
which can be allocated more sparsely and their presence depends heavily on the scene rather than a 
rather strict interval as in MPEG-4 Pt.\,2. 
This benefit is apparent when training a model to predict the start of a trim (see Tab.~\ref{tab:start_pred_arch}, top-right).
A potential solution would be to avoid \textit{inter}-frame codecs (\texttt{mpeg4} and H.264)
in favor of an \textit{intra}-frame codec (\eg MJPEG, ProRes).
However, this is a strong requirement for research datasets because it requires avoiding YouTube 
which stores videos compressed with inter-frame codecs (H.264 or VP9, according to view count).

\vspace{-1ex}
\paragraph{Reducing audio sampling rate.}
There is a substantial difference in the model's ability to predict  the start of a trim depending 
on the sampling rate of the audio track (Fig.~\ref{tab:start_pred_arch}, bottom-right).
While the reason behind the temporal artefacts in AAC is unknown, we recommend avoiding higher sampling rates. 
In our experiments, we rely on a 16kHz sampling rate as it provides a reasonable trade-off between audio quality and 
the start prediction performance.
Ultimately one would want to have a dataset with lossless audio tracks yet, again, it is a strong requirement for a dataset
as it is commonly used by YouTube.

\section{Experiments}\label{sec:experiments}

\paragraph{Dense in time dataset.}
The dataset is Lip Reading Sentences (LRS3)~\cite{afouras2018lrs3} which is obtained from TED talks for many speakers.
We use two variations of the dataset.
The first employs strict rectangular face crop coordinates that are extended to make a square (`dense in time and space').
The second variation consists of full-frame videos without cropping (`spatially sparse and temporally dense').
The raw videos are obtained from YouTube with RGB (25fps, H.264) and audio (16kHz, AAC) streams
and referred to as `LRS3-H264' and `LRS3-H264 (``No face crop'')'.
We utilise the \texttt{pretrain} subset and split video ids into 8:1:1 parts for train, validation, and test sets.
Only videos longer than 9 sec are used to unify it with the sparse dataset (discussed next).
In total, we use $\sim$58k clips from $\sim$4.8k videos. 

\vspace{-1ex}
\paragraph{Sparse in time dataset.} 
The dataset uses VGGSound \cite{Chen20a} which consists of 10s clips  collected from YouTube for 309 sound classes. 
A subset of `temporally sparse' classes is selected using the following procedure:
5--15 videos are randomly picked  from each of the 309 VGGSound classes, and manually annotated as to whether
audio-visual cues are only sparsely available.
After this procedure, 12 classes are selected ($\sim$4\,\%) or 6.5k and 0.6k videos in the train and test sets, respectively 
(for class names see Fig.\,\ref{fig:perf_per_class}).
Next, the second round of manual verification of a different subset of 20 videos from each class determines  
if it is feasible to align the sound based on the visual content. 
It is observed that $\sim$70\,\% of these video clips are synchronisable. 
We refer to this dataset as \textit{VGGSound-Sparse}.

\vspace{-1ex}
\paragraph{Baseline.}
Drawing on architectural details proposed in \cite{chen2021audio}, 
we design a baseline as a transformer decoder that uses audio features as \textit{queries} and 
visual features as context (\textit{keys} and \textit{values}) to predict the offset.
The audio features are pooled across the spectrogram frequency dimension and trained from scratch.
Apart from that, the feature extractors resemble ours.

\vspace{-1ex}
\paragraph{Offset grid.}
We define the synchronisation task as classifying the offset on a 
21-class grid ranging from $-$2 to $+$2 seconds with the 0.2-sec step size,
as explained in Sec.~\ref{sec:architecture}.
This can be regarded as a more challenging variant of the sync/off-sync task that prior work solves. 
We also experiment with a simpler setting with only 3 offset classes [$-1$, 0, $+$1], 
that test if a model could predict if one track either lags, is in sync, or is ahead of the other one. 

\vspace{-1ex}
\paragraph{Metrics.}
Considering the human off-sync perception tolerance,
in our experiments, we mainly report the Top-1 Accuracy with a $\pm$1 class of temporal 
tolerance~(as described in Sec.~\ref{sec:artefacts}).
Note, that the training loss does not account for this tolerance.
In the supplementary section, we additionally provide performance on accuracy without tolerance.

\subsection{Results}

\begin{table}
    \footnotesize
    \centering
    \begin{tabular}{l cc | cc | cc}
        \toprule
         & \multicolumn{2}{c|}{\textbf{Dense-Dense}} & \multicolumn{2}{c|}{\textbf{Dense-Sparse}} & \multicolumn{2}{c}{\textbf{Sparse-Sparse}} \\
         & \multicolumn{2}{c|}{\textit{LRS3 (Face crop)}} & \multicolumn{2}{c|}{\textit{LRS3 (W/o face crop)}} & \multicolumn{2}{c}{\textit{VGGSound-Sparse}} \\
         & 3 cls & 21 cls & 3 cls & 21 cls & 3 cls$^\star$ & 21 cls \\
        \midrule
        AVST$_\text{dec}$ & 98.4 & 89.8 & 95.8 & 83.1 & 52.2 & 29.3 \\
        Ours & 96.4 & 95.6 & 95.5 & 96.9 & 60.3 & 44.3 \\
        \bottomrule
    \end{tabular} 
    \vspace{1ex}
    \caption{\normalsize
        \textbf{The proposed model handles the increasing complexity of the setting and dataset better than the baseline
        while reaching a strong performance compared to the oracle.} 
        `Dense-Dense' refers to the face-cropped speech videos (LRS3), 
        `Dense-Sparse' for spatially-sparse LRS3 (`No face crop'),
        `Sparse-Sparse' is reported on VGGSound-Sparse which is sparse in time and space, \eg lion
        roars once during a clip.
        The synchronisation performance is measured in two settings: 
        the 3-class with ($-$1, 0, $+$1) offsets given 5-sec clips,
        and 21 classes of offsets from $-2.0$ to $+2.0$ sec with 0.2-sec step size.
        The latter setting allows $\pm$1 temporal class tolerance ($\pm$0.2 sec).
        $^\star$: oracle performance is 70\,\%.
    }
    \vspace{-2.0ex}
    \label{tab:main}
\end{table}
\begin{figure}
    \centering
    \includegraphics[width=\textwidth]{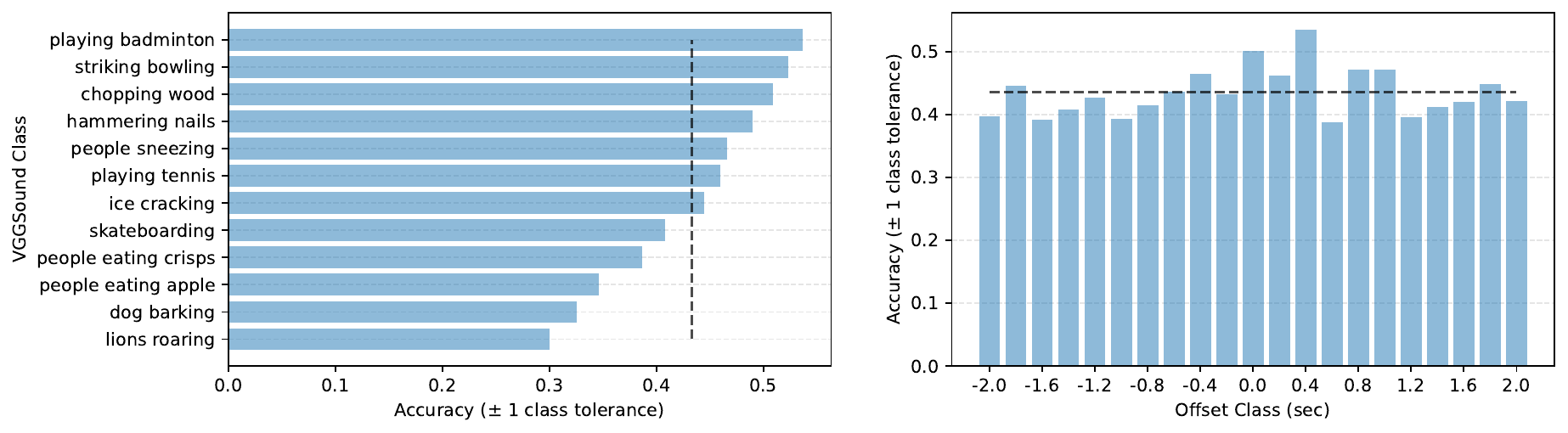}
    \vspace{-5ex}
    \caption{
        \textbf{Performance per data and offset class on VGGSound-Sparse (test).} 
    }
    \vspace{-3ex}
    \label{fig:perf_per_class}
\end{figure}

\paragraph{Dense in time and space.} 
Tab.~\ref{tab:main} shows the comparison between the baseline and proposed architecture.
As the task becomes more difficult (more sparse data or finer offset grid), 
we observe a larger gap between our model and the baseline.
In particular, the baseline performs strongly on LRS3 (dense in time and space) in the setting with
just three classes (98.4\,\%).
However, once the task gets more challenging, \ie when training on finer offset shifts (21 class, 0.2 sec apart),
the baseline performance deteriorates significantly ($\sim$89.8\,\%).
In contrast, the proposed model performs strongly even in the setting with finer offsets.
This suggests that the proposed model is better suited for more challenging data tasks.

\paragraph{Dense in time and sparse in space.}
A similar effect is observed on the dataset that is dense in time and sparse in space, \ie LRS3 (No face crop).
As both architectures drop their performance slightly on the 3-class setting after switching to a more difficult dataset,
the drop is more significant for the baseline than for our model.
Moreover, the baseline performance drops substantially in the 21-class setting ($>$6\,\%),
while our model performs strong.

\vspace{-2ex}
\paragraph{Sparse in time and space.}
Finally, the experiments on the VGGSound-Sparse reveal an even larger difference between the baseline method and 
our final model. 
For this experiment, we add additional data augmentation to mitigate overfitting.
In particular, our model significantly outperforms the baseline showing the benefit of
selectors on a more challenging dataset and setting. 
Ultimately, our model reaches 60\,\% in the 3-class setting which is close to 
the oracle performance ($\sim$70\,\%: a human performance on 240 randomly picked videos),
while achieving 44\,\% in the 21-class setting.
We report performance per class in Fig.\,\ref{fig:perf_per_class}.

\begin{table}
    \begin{minipage}[t]{0.60\textwidth}
        \vspace{0pt}
        \centering
        \footnotesize 
        \setlength\tabcolsep{0.40em}
        \begin{tabular}{cccc |c}
            \toprule
            & Sync Model & Pre-trained & Unfrozen & \\
            & pre-trained & Feature & Feature & ~ \\
            Selectors & on LRS3 & Extractors & Extractors &  Accuracy$_\text{21}$ \\
            \midrule
            \ding{55} & \ding{52} & \ding{52} & \ding{52} &  40.1 \\
            \ding{52} & \ding{55} & \ding{52} & \ding{52} &  12.1 \\
            \ding{52} & \ding{52} & \ding{55} & \ding{52} &  29.6 \\
            \ding{52} & \ding{52} & \ding{52} & \ding{55} &  33.5 \\
            \ding{52} & \ding{52} & \ding{52} & \ding{52} &  44.3 \\
            \bottomrule
        \end{tabular} 
        \vspace{2.3ex}
        \caption{\normalsize
            \textbf{Results of the ablation study.}
            The experiments are conducted on VGGSound-Sparse with 21 offset classes.
            Metric is Accuracy with $\pm$1 class tolerance.
        }
        \label{tab:abl_init}
    \end{minipage}\hfill
    \begin{minipage}[t]{0.37\textwidth}
        \vspace{-0.5ex}
        \centering
        \includegraphics[width=\textwidth]{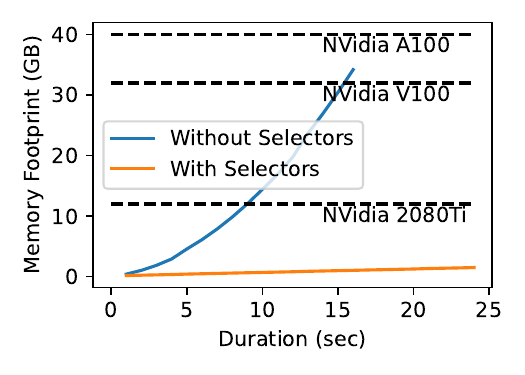}
        \vspace{-5ex}
        \captionof{figure}{\normalsize 
            Working with longer sequences quickly becomes infeasible without selectors.
        }
        \label{fig:mem_footprint}
    \end{minipage}
\end{table}

\begin{figure}
    \centering
    \vspace{-1ex}
    \includegraphics[width=1.0\textwidth]{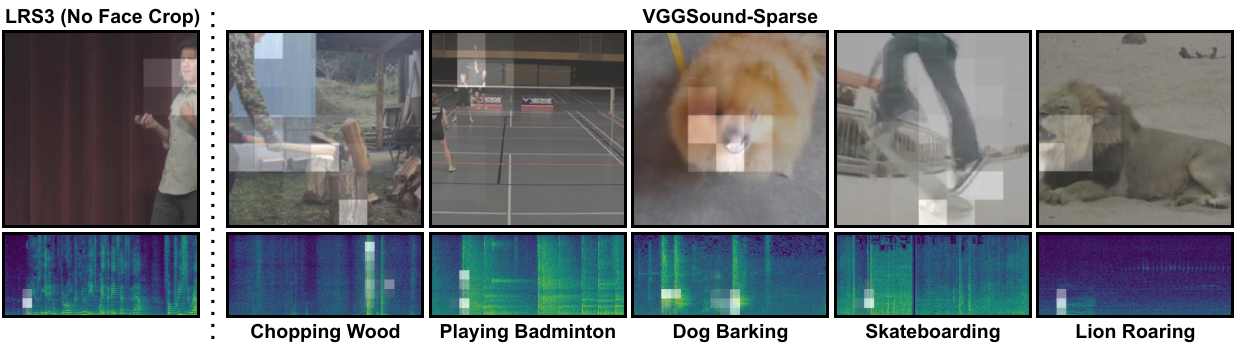}
    \vspace{-5ex}
    \caption{\normalsize 
        \textbf{Visual feature selectors focus on specific parts of the sparse signal that is useful for synchronisation.} 
        Examples are from the hold-out set of LRS3 (`No face crop') and VGGSound-Sparse.
        Attention is captured from a selector to a visual or spectrogram feature token from a head 
        within one of the layers. 
        Attention values are min-max scaled.
    }
    \label{fig:att_viz}
\end{figure}

\subsection{Ablation Study}\label{sec:abl_init}

In Tab.~\ref{tab:abl_init} we provide results for an ablation study.
The results are reported on the VGGSound-Sparse dataset with 21 offset classes.
More results are provided in the supplementary.

\vspace{-2ex}
\paragraph{Feature selectors.}
The architecture with feature selectors outperforms the vanilla transformer showing the effectiveness of selectors
in `compressing' signals from the audio and visual features.
Also, Fig.\,\ref{fig:mem_footprint} shows a memory footprint comparison of the 
two, omitting the memory consumed by feature extractors that are the same. 
It is evident that memory demand grows rapidly with the input duration making it impossible to work
with longer sequences and the transformer that inputs concatenated audio and visual streams.

\paragraph{Pre-training on dense signals.}
Pre-training a model on LRS3 (`No face crop') is an essential part of the training procedure on VGGSound-Sparse:
44.3 vs.~12.1 (near chance performance).
For this reason, we also pre-train the baseline architecture (see Tab.~\ref{tab:main}).

\vspace{-2ex}
\paragraph{Pre-trained feature extractors.}
The initialisation of audio and visual feature extractors with pre-trained weights has a strong positive effect on model performance.
To initialise our feature extractors, we use weights of S3D pre-trained on Kinetics 400 for action recognition 
and ResNet18 pre-trained sound classification on VGGSound.
The initialisation not only improves the final performance (43.3 vs.~33.5\,\%) but also significantly speeds up 
training.
We attribute this improvement to the fact that such initialisation allows the model to `skip' learning of the 
generic low-layer features and focus on training for synchronisation.

\vspace{-2ex}
\paragraph{Frozen feature extractors during training.}
Allowing the gradients to reach raw data pixels is useful for audio-visual synchronisation
as it makes the model sensitive to the smallest variations in the signal which is useful for synchronisation.
In particular, having feature extractors to be trainable significantly boosts the performance 
from 34 to above 43\,\%, and the difference is even more pronounced on LRS3 (`No face crop`) -- see supplementary material.

\vspace{-2ex}
\paragraph{Attention visualisation.}
Fig.~\ref{fig:att_viz} shows examples from LRS3-H.264 (`No Face Crop') and VGGSound-Sparse.
Specifically, the attention exhibits spatial locality as the selectors learned to attend to the features 
extracted from the mouth region as expected from a model trained on a speech dataset.
For a more challenging and diverse dataset, VGGSound-Sparse, the model highlights important parts of the visual and audio streams. 
In particular, the model accounts for the hit of the second badminton player who is far away in the background 
or attends to the axe during the chop, 
or the roaring mouth of the lion, yet these occur just once per video clip.
Similarly, audio feature selectors point to specific parts of the spectrogram when the change occurs.
More examples are provided in the supplementary material.

\begin{table}
\vspace{-1.5ex}
\begin{minipage}[c]{0.33\textwidth}
    \footnotesize 
    \setlength\tabcolsep{0.80em}
    \begin{tabular}{ccc}
        \toprule
        Length & \multicolumn{2}{c}{\textit{VGGSound-Sparse}} \\
        (sec.) & 3 classes & 21 classes \\
        \midrule
        2 & 55.6 & ~~---  \\
        3 & 59.4 & 36.8 \\
        4 & 60.8 & 43.0 \\
        5 & 60.3 & 44.2 \\
        6 & 61.2 & 45.6 \\
        7 & 62.9 & 46.5 \\
        \bottomrule
    \end{tabular} 
\end{minipage}\hfill
\begin{minipage}[c]{0.65\textwidth}
\vspace{2ex}
\caption{\normalsize \textbf{Synchronisation accuracy improves with input length.}
We report results on two settings: with 3 offset classes ($-$1, 0, $+$1 sec) and 
21 classes ($\pm$2.0 sec grid with 0.2-sec step size).
The results are reported on the test subset and accuracy is used as the metric.
The accuracy for the 21-class setting is reported with $\pm$1 class tolerance.
We use the same input lengths for pre-training, fine-tuning, and testing.
\label{tab:diff_lengths}}
\end{minipage}
\vspace{-3ex}
\end{table}

\vspace{-2ex}
\paragraph{Input length.}

\blue{
As the sparse synchronisation cues occur only occasionally within a video clip, processing shorter temporal crops decrease the 
chance of having sufficient cues for synchronisation, which, in turn, should decrease the performance.  
In Tab.~\ref{tab:diff_lengths}, we show how performance varies with respect to the duration of input video clips.
The results on the 3 and 21 offset classes illustrate the upward trend in model performance as the input duration extends.
Note that the longer the inputs, the less unseen training data the model processes at each epoch.
Specifically, a 10-second clip may be split into non-overlapping 3-second clips, which is not possible 
with clips longer than 5 seconds.
Thus, this effect may undermine the current performance.}
\section{Conclusion}
In this work, we study `in the wild' videos that often have a synchronisation signal that is sparse in time.
This requires a model to efficiently process longer input sequences as these synchronisation cues occur only rarely.
To this end, we designed a transformer-based synchronisation model that has linear complexity with respect to the input length.
This was made possible by using a small set of learnable `selectors' that summarise long audio and visual features that are employed 
to solve the synchronisation task.
To evaluate models in this challenging setup, we curate a dataset with only sparse events and train it on 5-second long clips.
Finally, we discovered that compression artefacts caused by audio and video codecs might pose a threat to
training for synchronisation, yet, as we show, these artefacts are easily identifiable and could be avoided to a certain extent.

\vspace{-1ex}
\blue{\paragraph{Limitations.}
First, considering the complicated input-output relationship in the proposed model, it is challenging to determine
which part of the input signal influences the output.
Second, in this work, we considered signals that are `dense in time and space', `dense in time but sparse in space',
and `sparse in time and space'. 
However, there is another interesting setting `sparse in time but dense in space' yet it is not clear how
to design such a dataset without making it `too artificial'.
Third, despite showing strong performance on the proposed dataset, VGGSound-Sparse, there is still room for improvement.
}

\vspace{-1ex}
\blue{\paragraph{Acknowledgements.} Funding for this research was provided by the
Academy of Finland projects 327910 and 324346,
EPSRC Programme Grant VisualAI EP$\slash$T028572$\slash$1, and a Royal Society Research Professorship.
We also acknowledge CSC -- IT Center for Science, Finland, for computational resources.}

\begin{figure}[t]
    \centering
    \includegraphics[width=\textwidth]{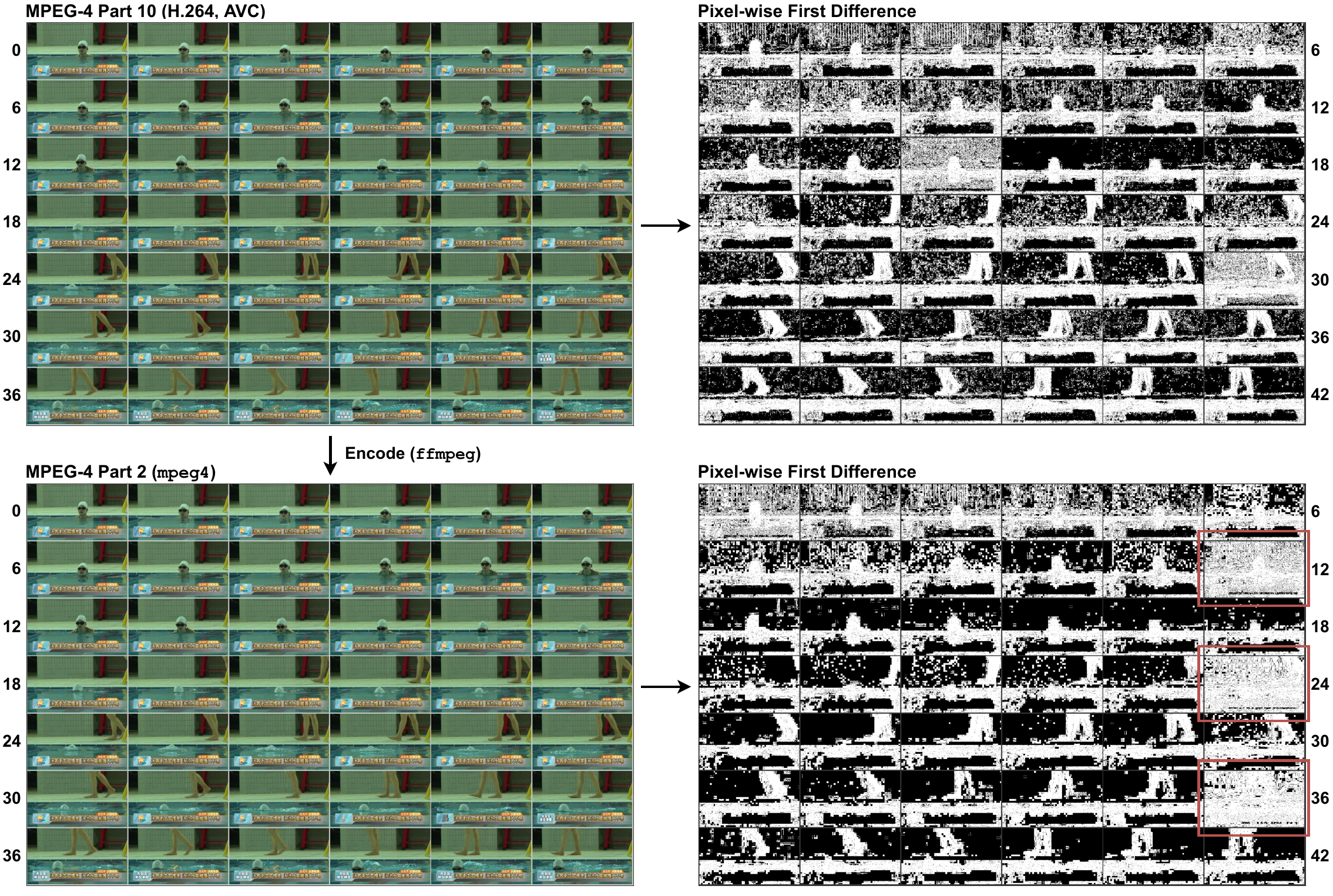}
    \setlength\fboxrule{1pt}
    \vspace{-4ex}
    \caption{\normalsize
        \textbf{MPEG-4 Part 2 (\texttt{mpeg4}) leaks regular temporal artefacts every \fcolorbox{brick}{white}{12$^\text{th}$ frame}
        that are unnoticeable on the RGB stream.} 
        The top row shows a video encoded with H.264 and the pixel-wise first difference between two consecutive frames.
        The bottom row shows the same video but now transcoded (encoded) from H.264 into MPEG-4 Pt.\,2 (\texttt{mpeg4})
        and its first difference.
        The first difference illustration is made sensitive to \textit{any} change in corresponding pixel intensity (colored in white).
    }
    \label{fig:rgb_diff}
\end{figure}

\section{Supplementary Material}

In this section, we present more details on temporal artefacts (Sec.~\ref{sec:time_art}), additional ablation study 
results (Sec.~\ref{sec:more_abl_res}), and implementation details (Sec.~\ref{sec:imp_details}).

\subsection{Further Discussion on Temporal Artefacts}\label{sec:time_art}

\paragraph{Illustrating temporal artefacts.}
The temporal artefacts in the audio and visual streams have a periodic nature, 
\eg artefacts are equidistant temporally, 
and can be easily highlighted on the RGB stream.
In Fig.~\ref{fig:rgb_diff}, we compare the first-order temporal difference calculated from a video that was 
encoded with MPEG-4 Part 10 (H.264 or AVC) and MPEG-4 Part 2 (aka \texttt{mpeg4}) which was 
transcoded\footnote{Sometimes this process is incorrectly called `re-formatting' or `re-encoding''}
from H.264.
Notice the white noise (RGB intensity change) on the 12$^\text{th}$, 24$^\text{th}$, 36$^\text{th}$ frames
(the period is 12) in the bottom-right image.
In particular, notice that the caption at the bottom of each frame is fixed across the video yet every 12$^\text{th}$ 
we see it to be white, and black otherwise.
The 12-frame period originates from the Group of Picture (GoP) size, \ie how often independently-encoded 
frames occur, and is selected by the \texttt{ffmpeg} tool by default.

In contrast, on a video encoded in H.264, these artefacts are also visible but are not regular
(see the top row of Fig.~\ref{fig:rgb_diff}) since the H.264 codec follows a more flexible 
key-framing algorithm that 
allows it to reference not 1 as in \texttt{mpeg4} but up to 16 different key-frames (I-frames).
Moreover, the allocation of key-frames depends heavily on the scene content, which makes the temporal 
spacing of key-frames unique for each video and, thus,  less predictable \textit{a priori}.
Note: this should hold for any video originally encoded with regularly spaced key-frames, \ie transcoding will not remove temporal artefacts.
For instance, if we crop out and transcode 5 seconds of a 10-second MPEG-4 Part~2 video file into H.264, 
the artefacts will still be present in the RGB stream.

\vspace{-3ex}
\paragraph{Tracking accuracy metrics with temporal tolerance.}
We noticed that, when the model is trained for start-time prediction
and there are temporal artefacts in the data, the model
tend to output equidistant classes in the top-K predictions. 
For example, if the correct start class is 15 (15$^\text{th}$ frame if a video is encoded at 10 fps), 
the model would be `confident' in classes (3, 15, 27, etc) 
or simply any class where `\textit{class id} $\text{mod}$ \textbf{12} = 3'.
The 12 results from the GoP size as described in the previous paragraph.
This differs from the expected error behavior where predictions would be around the true class, 
\eg (13, 15, 14, etc) or 1.3sec $\pm$0.1sec.

\begin{figure}
    \centering
    \includegraphics[width=\textwidth]{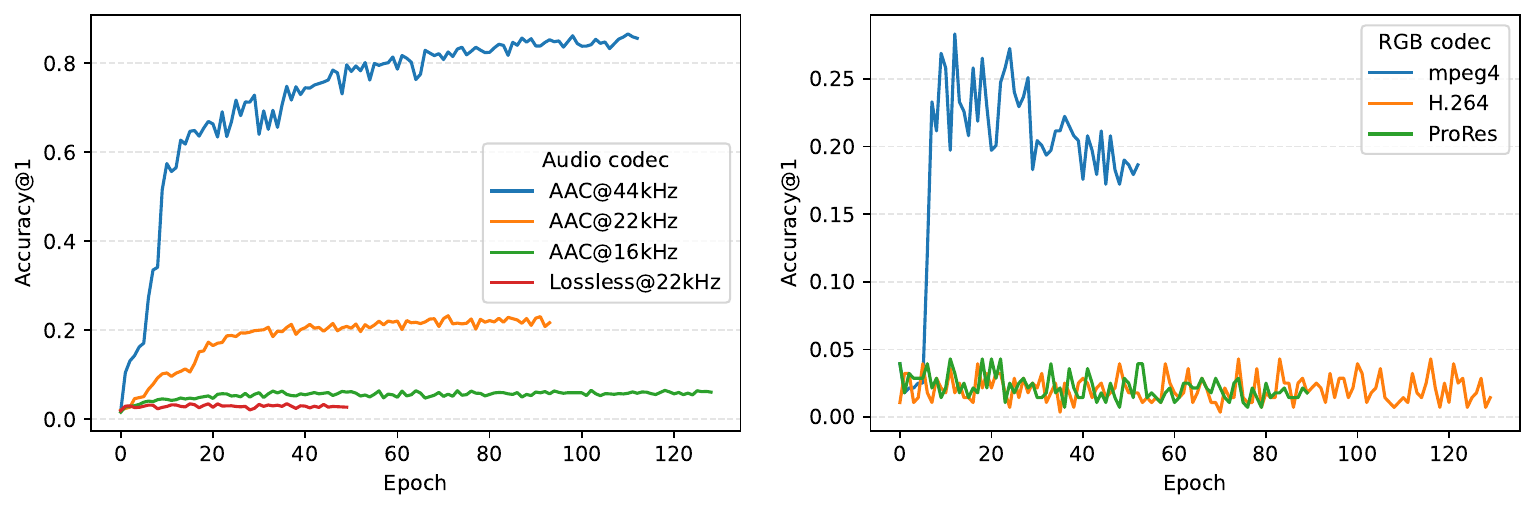}
    \caption{\normalsize
        \textbf{Some codecs leak temporal artefacts and it is easy to notice when training for the temporal crop start-time prediction.} 
        Training dynamics for the start-time prediction experiments.
        Left: random noise encoded in AAC with varying sampling rates.
        Right: ProRes videos encoded as H.264 or \texttt{mpeg4}.
        Metric is the top-1 accuracy across 50 classes.
    }
    \vspace{-3ex}
    \label{fig:training_curve}
\end{figure}

\vspace{-3ex}
\paragraph{Training details.}
The artefact leakage can be detected by training a model to predict the start-time of a temporal segment 
given a 5-second audio spectrogram or a stack of RGB frames.
In our experiments, we used a 50-way classification on a grid with the 0.1s (one frame) stride from a 10fps video. 
The 50 classes is because in our implementation the start of the 
temporal crop could only begin during the first 5 seconds.
For simplicity, we used a ResNet18 followed by a spatial average pooling and a dense layer as a classification head for both audio 
and visual inputs.
The training dynamic is shown in Fig.~\ref{fig:training_curve}.

\subsection{Ablation Study (Additional Results)}\label{sec:more_abl_res}

\begin{table}
    \centering
    \small
    \setlength\tabcolsep{0.60em}
    \begin{tabular}{cccc|cc}
        \toprule
        & Sync Model & Pre-trained & Unfrozen & \textbf{Dense-Sparse} & \textbf{Sparse-Sparse} \\
        & pre-trained & Feature & Feature & \textit{LRS3 (No face crop)} & \textit{VGGSound-Sparse} \\
        Selectors & on LRS3 & Extractors & Extractors &  Acc$_\text{21}$ / Acc$^\text{tol.}_\text{21}$ & Acc$_\text{21}$ / Acc$^\text{tol.}_\text{21}$ \\
        \midrule
        \ding{55} & \ding{52} & \ding{52} & \ding{52} & 78.3 / 97.2 & 21.2 / 40.1 \\
        \ding{52} & \ding{55} & \ding{52} & \ding{52} & ---         &  5.0 / 12.1 \\
        \ding{52} & \ding{52} & \ding{55} & \ding{52} & 76.6 / 94.1 & 16.0 / 29.6 \\
        \ding{52} & \ding{52} & \ding{52} & \ding{55} & 38.6 / 69.7 & 14.7 / 33.5 \\
        \ding{52} & \ding{52} & \ding{52} & \ding{52} & 80.7 / 96.9 & 26.7 / 44.3 \\
        \bottomrule
    \end{tabular} 
    \vspace{1ex}
    \caption{\normalsize
        \textbf{Additional results of the ablation study}.
        The metrics are the accuracy with and without temporal tolerance  ($\pm$1 class) on a 21-class offset grid ranging 
        from $-$2.0 sec to $+$2.0 sec with a step size of 0.2 sec.
    }
    \label{tab:abl_more}
\end{table}

\begin{table}
    \begin{minipage}[c]{0.48\textwidth}
        \centering
        \small
        \begin{tabular}{c c c}
            \toprule
             & & \textit{LRS3 (No face crop)} \\
            Depth & Params. & Acc$_\text{21}$ / Acc$^\text{tol.}_\text{21}$ \\
            \midrule
            1 & 31.2M & 61.7 / 89.2  \\
            2 & 42.8M & 78.0 / 95.7  \\
            3 & 55.3M & 80.7 / 96.9  \\
            6 & 89.0M & 81.1 / 97.5  \\
            \bottomrule
        \end{tabular} 
    \end{minipage}\hfill
    \begin{minipage}[c]{0.51\textwidth}
        \caption{\normalsize
            \textbf{The performance scales with the depth of the transformer modules and saturates by the 3-layer capacity.}
            The results are reported on the setting with 21 classes with offsets from $-2.0$ to $+2.0$ with step size 0.2 sec.
            Metrics are accuracy with and without temporal tolerance ($\pm$1 class).
        }
        \label{tab:abl_layers}
    \end{minipage}
\end{table}

As mentioned in  Sec.~\ref{sec:experiments}, we report additional results with metrics without temporal
tolerance as well as the performance on LRS3-H.264 (`No face crop') in Tab.~\ref{tab:abl_more}.
Moreover, we experiment with different numbers of layers in the transformer modules, 
provide additional visualisation of attention, and show the benefit of fine-tuning the model on a 
more diverse and large-scale dataset.

\paragraph{Results with additional metrics and a dataset.}
Similar to the results on the VGGSound-Sparse dataset,
we see the importance of each component in the final model performance when
compared on the `Dense-Sparse' setting, \ie LRS3-H.264 (`No Face Crop').
Notice, the narrower gap in performance compared to the models with and without selectors.
This highlights that the use of selectors becomes more important as the complexity of the task increases (from `Dense' to `Sparse').

Interestingly, the feature extractors pre-training has a much more noticeable effect on the VGGSound-Sparse compared with
the LRS3 (`No face crop') dataset.
We attribute it to the simplicity of the talking-face dataset, which imposes lower requirements on the diversity of audio-visual 
synchronisation cues.
Such cues are easier and faster to learn because they are quite generic across videos.

\vspace{-3ex}
\paragraph{Number of transformer layers.}
Tab.~\ref{tab:abl_layers} presents the impact of the transformer's depth on the model's synchronisation performance.
We vary the depths of two feature selector transformers and the synchronisation transformer simultaneously.
According to the results,  the performance grows as the capacity of the model increases, however, the returns are diminishing.
Thus, we regard the 3-layer option as our final model.

\begin{figure}[t]
    \centering
    \includegraphics[width=1.0\textwidth]{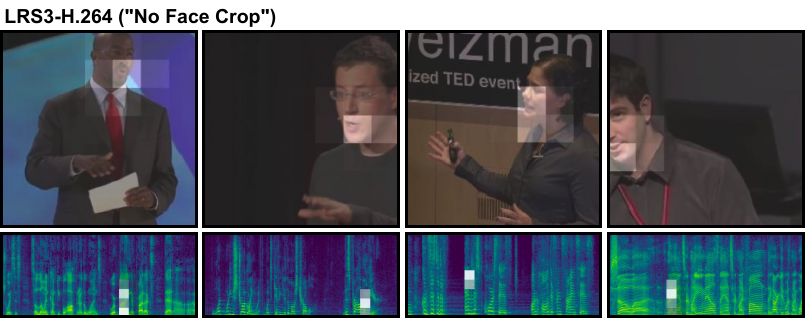}
    \vspace{-5ex}
    \caption{\normalsize
        \textbf{Visual feature selectors focus on specific parts of the sparse signal that is useful for synchronisation.} 
        Examples are from the hold-out set of LRS3 (`No face crop').
        Attention is captured from a selector to a visual or spectrogram feature token from a head 
        within one of the layers. Attention values are min-max scaled.
    }
    \vspace{-1ex}
    \label{fig:att_viz_sup}
\end{figure}

\vspace{-3ex}
\paragraph{Additional attention visualisations.}
We provide additional attention visualisation on the LRS3-H.264 (`No Face Crop') dataset in Fig.~\ref{fig:att_viz_sup}.
The illustration shows that the model performs well regardless of gender, race, partial occlusion, 
position of a head within a frame, as well as the different distance to a person, and background.

\begin{table}
    \centering
    \small
    \begin{tabular}{llc}
        \toprule
        Pre-trained on & Fine-tuned on & Acc$_\text{21}$ / Acc$^\text{tol.}_\text{21}$ \\
        \midrule
        LRS3 (`No Face Crop') & VGGSound-Sparse & 26.7 / 44.3 \\
        LRS3 (`No Face Crop') & VGGSound & 33.5 / 51.2 \\
        \bottomrule
    \end{tabular} 
    \vspace{1ex}
    \caption{\normalsize
        \textbf{Fine-tuning on a larger general-purpose dataset significantly improves synchronisation performance}.
        The results are shown on the test-set of VGGSound-Sparse.
        Metrics are accuracy with and without temporal tolerance ($\pm$1 class).
    }
    \label{tab:vggsound_full}
\end{table}

\vspace{-3ex}
\paragraph{Fine-tuning on a larger dataset.}

We found that fine-tuning on a larger general-purpose dataset, 
\eg VGGSound (300+ classes), gives a significant boost to 
synchronisation performance of the sparse classes. 
The results of this experiment are shown in Tab.~\ref{tab:vggsound_full}.

\subsection{Implementation Details}\label{sec:imp_details}

\paragraph{Feature extractors.}
For visual input, we sample the video with 25fps and resize the frames such that $\min(H_v\comma W_v)=256$ 
and take a random crop of $ 224^2 $.
The 5-second stack of cropped frames ($125 \times 224 \times 224 \times 3 $) 
is encoded with S3D which results in a $ 16 \times 7 \times 7 \times 1024 $ visual feature map.
For the 5-second audio input, we convert a waveform sampled at 16kHz to a spectrogram with the STFT (512 in size with 128 hop length) 
and apply a logarithm to the result. 
The log-spectrogram ($257\times 626$ is passed to the ResNet18, which outputs an audio feature map of size $9 \times 20$.

\vspace{-2ex}
\paragraph{Feature selectors and audio-visual synchronisation transformer.}
Before processing the visual feature map in the feature selector, a simple $1\times1$ convolution layer 
is used to map the visual features from $1024$-d to $512$-d which is the dimension of the transformer blocks across the final 
architecture. 
We use layer norm~\cite{ba2016layer} on both audio and visual features and apply positional encoding after that.
Both feature selectors have the same number of trainable `selectors'~($k_a = k_v = 16$)
that are initialised with Gaussian noise.
As has been discussed in the ablation study, 3 cross-attention layers are used for each feature selector and 
3 attention layers in the audio-visual synchronisation transformer.
The layers have 8 heads.

\vspace{-2ex}
\paragraph{Data augmentations.}
We use two sets of augmentations for (pre-)training on LRS3 and finetuning on VGGSound-Sparse.
For the RGB stream, we use: random spatio-temporal 
crop\footnote{The range of the temporal crop is limited such that the resulting clip could fit the
audio that was offset as well as the visual clip.} and random horizontal flip (with 50\,\% probability).
For VGGSound-Sparse, we additionally may crop smaller to $192^2$ (with 50\,\% probability) followed by upscaling to $224^2$ 
as well as applying colour jittering and converting to a grey scale (with 20\,\% probability and only for VGGSound-Sparse).

For the audio, we found that adding short audio jittering around the start of the offset audio trim 
(\eg $\sim U[-0.05\comma +0.05]$ seconds) stabilises training.
Additionally, for VGGSound-Sparse, we apply (with 20\,\% probability) random reverberation, amplitude jittering, random pitch shift, 
a low-pass filter, and add small Gaussian noise. 
We also employ \textit{SpecAugment} \cite{park2019specaugment} 
except for time stretching as it could potentially mislead the synchronisation signal.

\vspace{-2ex}
\paragraph{Training details.}
The final synchronisation model is trained for 8 days on 8 NVidia V100 GPU (32Gb) with half-precision on LRS3-H.264
followed by a few hours of finetuning on VGGSound-Sparse.
During finetuning, we use the same hyper-parameters as for pre-training (except for the data augmentation as mentioned earlier).
The models are trained with a batch size of 10 video clips per GPU with a learning rate of $5\cdot10^{-6}$ scaled by
the number of GPUs.
We warm up training with a 100-times lower learning rate and linearly increase it during the first 1k iterations.
During training, 
Adam optimizer is used with $\beta_{1,2} = [0.99, 0.999]$ and 
the norm of the gradients is max-clipped to 1.
The code base relies on \texttt{PyTorch} (including \texttt{torchvision} and \texttt{torchaudio}) 
\cite{paszke2019pytorch} and \texttt{ffmpeg} (version 4.3.2).
Before training, we fix the set of offsets for every video in the validation and test sets such that 
we could evaluate using the same data during experimentation.

\bibliography{egbib}
\end{document}